%% file: main.tex
\documentclass[letterpaper, 10 pt, conference]{ieeeconf}  

\usepackage{siunitx}
\usepackage{multirow}
\usepackage[matmatrix,opt]{icpslab}
\usepackage{algorithm}
\usepackage{algpseudocode}
\usepackage{array}
\IEEEoverridecommandlockouts                              
\usepackage{hyperref}
\usepackage{graphics} 
\usepackage{epsfig} 
\usepackage{times} 
\usepackage{amsmath} 
\usepackage{amssymb}  
\usepackage{cite}
\usepackage{comment}
\usepackage[dvipsnames]{xcolor}

\newtheorem{remark}{Remark}
\newcommand{\MPCC}{\pi_q}
\newcommand{\col}{\mathrm{col}}
\newcommand{\etabold}{\boldsymbol{\eta}}
\newcommand{\thetabold}{\boldsymbol{\theta}}
\newcommand{\lambdabold}{\boldsymbol{\lambda}}
\newcommand{\betabold}{\boldsymbol{\beta}}

\newcommand{\pdiff}[2]{\frac{\partial #1}{\partial #2}}

\newcolumntype{C}[1]{>{\centering\arraybackslash}p{#1}}

\usepackage{todonotes}
\newcommand{\Nam}[1]{\textcolor{purple}{#1}}

\newcommand{\tcb}[1]{\textcolor{blue}{#1}}

\newif\iffinal
\finaltrue   

\iffinal
  \renewcommand{\Nam}[1]{#1}
  
  \renewcommand{\tcb}[1]{#1}

\fi

\title{AD-MPCC: Adaptive Differentiable Model Predictive Contouring Control \\ for
Autonomous Racing}
\author{Authors\thanks{Thanks editors and reviewers.    }}
\author{Nam T. Nguyen$^{1, \dagger}$, Binh Nguyen$^{1, \dagger}$, Ahmad Amine$^{2}$, \\Thanh Vo-Duy$^{3}$, Rahul Mangharam$^{2}$, and Truong X. Nghiem$^{1}$
\thanks{This material is based upon work supported by the National Science Foundation under Award No. 2514584.}
\thanks{$^\dagger$These authors contributed equally to this work.}
\thanks{$^{1}$Department of Electrical and Computer Engineering, University of Central Florida, Orlando, FL 32816, USA.}%
\thanks{$^{2}$Department of Electrical and Systems Engineering, University of Pennsylvania, Philadelphia, PA 19104, USA.}%
\thanks{$^{3}$CTI Lab4EV, School of Electrical and Electronic Engineering, Hanoi University of Science and Technology, Hanoi 100000, Vietnam.}%
}

\begin{document}
\maketitle
\thispagestyle{empty}
\pagestyle{empty}

\setlength{\abovedisplayskip}{3pt}
\setlength{\belowdisplayskip}{3pt}
\setlength{\textfloatsep}{8pt}
\setlength{\floatsep}{6pt}
\setlength{\intextsep}{6pt}
\begin{abstract}
This paper presents Adaptive Differentiable Model Predictive Contouring Control (AD-MPCC), a framework for autonomous racing that integrates differentiable MPCC with online parameter estimation to handle varying road-surface conditions. For online parameter estimation, we leverage a parameterized Pacejka Magic Formula together with a regularized moving-horizon estimation scheme with exponentially decaying weights to capture road interactions and update parameters in real time.  Furthermore, we propose a differentiable MPCC (Diff-MPCC) framework that enables optimal adjustment of objective weights based on predefined long-horizon performance costs.
To implement Diff-MPCC for online objective weight adaptation, we propose a Pacejka-informed machine learning model that is trained in a supervised manner using data generated by Diff-MPCC to tune the objective weights.
Simulation results demonstrate that AD-MPCC reliably ensures safety and achieves faster lap times compared to  baseline controllers in both single-surface and  multiple-surface scenarios.
\end{abstract}





\input{sec_intro}
\input{sec_problem}
\input{MPCC}
\input{sec_method_2}
\input{adaptation}

\input{diffMPCC}

\input{sec_results}

\input{sec_conclusion}





\bibliographystyle{IEEEtran}
\bibliography{references}

\end{document}

%% file: sec_intro.tex
\section{Introduction}
\label{sec:intro}
Autonomous racing pushes vehicles to operate at their physical limits while optimizing well-defined performance objectives such as lap-time minimization \cite{betz2022autonomous}, making it a compelling testbed for advanced control methods, including imitation learning \cite{10186780}, reinforcement learning \cite{10982032}, \Nam{or} iterative learning control \cite{11260933}. Model predictive contouring control (MPCC) \cite{lam_model_2010, liniger2015optimization} has emerged as a standard control framework for autonomous racing. By jointly minimizing contouring and lag errors while maximizing progress along a reference path, MPCC naturally captures the trade-off between tracking accuracy and racing speed.

A growing body of work has sought to improve MPCC by replacing or augmenting the physics-based predictive model with data-driven models.
Gaussian process models have been integrated with nominal dynamics in \cite{kabzan2019learning, pinho2023learning}, and neural networks have been used to compensate for unmodeled effects in \cite{gomes2024learning}, with both approaches demonstrating improved lap times over their physics-based counterparts \cite{hewing2020learning}. 
However, the nonlinearity and nonconvexity of the learned components increase the complexity of the predictive model, complicating real-time deployment.
In practice, the dominant source of model uncertainty in racing arises from tire-road interactions, which are well characterized by the Pacejka tire model \cite{pacejka2005tire}. 
Rather than introducing learned residual models, directly estimating Pacejka parameters preserves \Nam{simplicity} and the interpretability of the vehicle dynamics based on physics.  
On multi-surface tracks, Pacejka parameters must be estimated in real time  to ensure the predictive model's accuracy.

In addition to online parameter estimation, another challenge in autonomous racing on multi-surface tracks is that changes in the dynamic model caused by updates of the Pacejka parameters require corresponding adaptation of the MPCC objective weights to maintain safety and consistent control performance.
Differentiable MPC provides a principled mechanism for optimizing MPC objective weights by differentiating a performance cost through the control optimization problem, enabling gradient-based weight updates \cite{11373898, DiffMPC}.
Recent work has further enabled GPU-accelerated differentiable MPC solvers that scale to challenging driving tasks \cite{adabagDifferentiableModel2025}. Extending this idea to MPCC, however, faces a fundamental difficulty.
In standard MPC, the reference trajectory is fixed and the differentiation is straightforward. In MPCC, however, the reference point is determined implicitly through the optimized progress variable $\theta$, coupling the objective weights to the reference trajectory and requiring higher-order derivatives of the reference path \cite{lam_model_2010}.

\begin{figure}[!t]
    \centering
    \includegraphics[trim={0.0cm 0 0 0},clip, width=0.97\linewidth]{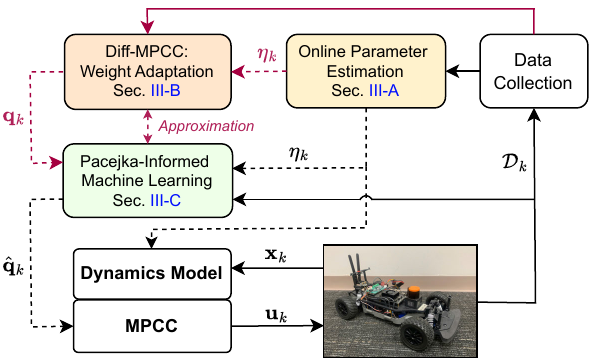}
    \vspace{-5pt}
    \caption{The AD-MPCC architecture for autonomous racing. Purple lines and black lines represent offline training and real-time deployment, respectively. Dashed lines denote parameters and solid lines denote signals.}
    \label{fig:adap_MPCC}
\end{figure}

Addressing the above challenges, this paper presents the Adaptive Differentiable MPCC (AD-MPCC) framework that combines online Pacejka parameter estimation with a differentiable MPCC (Diff-MPCC) for autonomous racing under different road-surface conditions. Diff-MPCC provides gradients of the MPCC solution with respect to the objective weights through the implicit optimality conditions of the underlying nonlinear program, enabling their optimization via a long-horizon performance cost. 
For online parameter estimation, we formulate a prior-regularized moving horizon estimation \Nam{(MHE)} problem with exponentially decaying weights that prioritizes recent observations to capture surface transitions.
AD-MPCC integrates these two components to update the predictive model and tune the controller weights at each time step.
Fig.~\ref{fig:adap_MPCC} illustrates the AD-MPCC architecture.

Due to the computational burden of Diff-MPCC, we further propose a Pacejka-informed machine learning (PaIML) technique to efficiently approximate the optimal MPCC weight solutions obtained from Diff-MPCC. 
Specifically, PaIML reduces the input dimension of the online weight adaptation model from 14 to 5 by leveraging the Pacejka tire model and the vehicle’s progress along the racetrack. 
The PaIML model is trained using data generated by Diff-MPCC, collected by running the vehicle on the racetrack under varying conditions. 
The trained model is then integrated into the AD-MPCC framework to infer near-optimal MPCC weights from the estimated vehicle dynamics in real time.

\Nam{To highlight the performance of AD-MPCC, we validate the proposed framework in the F1TENTH-Gym environment \cite{o2020f1tenth} across both single- and multi-surface scenarios. The results show that AD-MPCC and Diff-MPCC significantly reduce lap times on a single-surface track, while AD-MPCC is the only controller capable of completing laps on an multi-surface road, thereby ensuring safe vehicle operation during racing in an uncertain environment.}

The contributions of this work are highlighted as follows.
\begin{itemize}
    \item We propose a new AD-MPCC framework that performs online estimation model and objective weight adaptation for autonomous racing.
    \item We provide a prior-regularized MHE scheme for online estimation that rapidly estimate Pacejka parameters in the dynamic model. 
    \item
    We develop a differentiable programming technique for MPCC that enables gradient-based weight tuning to improve performance of the racing car.
    Also, a real-time version of Diff-MPCC is designed by proposing a Pacejka-informed ML technique.
    
\end{itemize}


\textit{Notation:} Let $\mathbb{R}$, $\bbN$, $\mathbb{R}_{>0}$, and $\mathbb{N}_{>0}$ denote the set of real numbers, integer number, positive real numbers, and positive integer number, respectively.
For vectors $\bfv_1, \dots, \bfv_n$, denote $\col(\bfv_1,\dots,\bfv_n) = \mm{\bfv_1^\top, \dots, \bfv_n^\top}^\top$.
For  vector $\bfu=\mm{u_1, \dots, u_n}^\top \in \bbR^n$, let $\diag({\bfu})$ be the diagonal matrix whose diagonal is $\bfu$.
For differentiable functions $f(x): \bbR^m \rightarrow \bbR^n$ and $g(z): \bbR^n \rightarrow \bbR$, $\partial_x f$ is the Jacobian matrix of $f$ and $\nabla_{zz}^2 g$ is the Hessian matrix of $g$.

%% file: sec_problem.tex
\section{Preliminaries}
\label{sec:problem}


\subsection{Dynamic vehicle model}

\begin{figure}[t]
    \centering
 \includegraphics[trim={0.3cm 0.1cm 0.3cm 0.1cm},clip, width=0.7\linewidth]{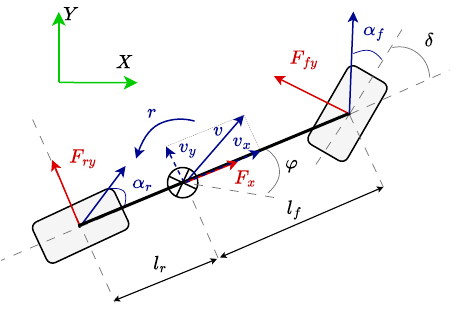}
 \vspace{-2pt}
    \caption{Single-track dynamic bicycle model of a 
   car.}
    \label{fig:car}
\end{figure}

This paper considers the impact of track/road surface condition on the dynamics of the vehicle. 
These effects manifest primarily through lateral and longitudinal forces acting on the vehicle.
Thus, we adopt a dynamic bicycle model \cite{kabzan2019learning}, which provides a compact representation of these effects
and is depicted in Fig.~\ref{fig:car} with the vehicle states and acting forces shown.
Here, $\mathbf{x} =  \col(X, Y, \varphi, v_x, v_y, r, \delta)$ denotes the state of the car, where 
$X$ and $Y$ are the positions in the global Cartesian coordinate frame, 
$\varphi$ represents the heading angle, 
$v_x$, $v_y$, $r$, and $\delta$ stand for the longitudinal velocity, the lateral velocity, the yaw rate, and the steering angle, respectively.
The control input vector is $\mathbf{u} = \mm{\Delta_\delta, \Nam{F_a}}^\top$, where $\Nam{F_a}$ is the motor driving force and $\Delta_\delta$ is the steering angle speed.




The dynamic bicycle model is given by
\begin{equation}
    \dot{\mathbf{x}} = f(\mathbf{x},\mathbf{u}) =
    \begin{bmatrix}
        v_x\cos(\varphi) - v_y\sin(\varphi) \\
        v_x\sin(\varphi) + v_y \cos(\varphi) \\
        r \\
        \frac{1}{m}(F_x - F_{f,y} \sin(\delta) +mv_yr) \\
        \frac{1}{m}(F_{r,y} + F_{f,y} \cos(\delta) - mv_xr) \\
        \frac{1}{I_z} (F_{f,y}l_f\cos(\delta) - F_{r,y} l_r)  
        \\
        \Delta_\delta
    \end{bmatrix} \text,
    \label{eq:dyna_model_1}
\end{equation}
where $l_r$ and $l_f$ are the distances from the center of gravity (CoG) of the car to the rear and front wheels, $I_z$ is the yaw inertia of the car, and $m$ is the mass of the car.
The lateral forces for the front $F_{f,y}$ and rear $F_{r,y}$ wheels are given by
the Pacejka model \cite{pacejka2005tire} 
\begin{align}
    &\mu_r = D_r \sin\!\left(C_r \arctan\!\left(B_r \alpha_r\right)\right), \label{eq:mu_r_y}\\
    &\mu_f = D_f \sin\!\left(C_f \arctan\!\left(B_f \alpha_f\right)\right), \label{eq:mu_f_y}\\
    &\alpha_r = \arctan\!\left(\frac{v_y - l_r r}{v_x}\right)\!\text, \;
    \alpha_f = \arctan\!\left(\frac{v_y + l_f r}{v_x}\right) \!-\! \delta, \notag\\
    &F_{r,y} = \frac{l_f}{l_r + l_f} m g \mu_r, \quad
    F_{f,y} = \frac{l_r}{l_r + l_f} m g \mu_f. \label{eq:force_y}
\end{align}
where $D_r, C_r$, $B_r$,  $D_f, C_f$, and $B_f$ are positive scalar parameters, and $g$ is the gravitational constant.
Here, $\mu_r$ and $\mu_f$ represent the friction coefficients between the rear tire and the front tire to the road surface, with $\mu_r, \mu_f \in [-\mu_\textrm{max}, \mu_\textrm{max}]$, where $\mu_\textrm{max}$ is the maximum value of the Pacejka line with respect to the type of road \cite{pacejka2005tire}.
Additionally, the forces $F_{f,y}$ and $F_{r,y}$ are formulated with the assumption that there are no lateral acceleration effects. 
$\alpha_f$ and $\alpha_r$ are the side-slip angles of the front and rear wheels, respectively.

The longitudinal force $F_x$ is generated by the rear-wheel drivetrain and provides the acceleration thrust for the car \cite{kabzan2019learning}. 
It is approximated by a linear combination of the motor driving force and resistances
\begin{align}
    F_x = C_a \Nam{F_{a}} - C_{r0} - C_{r2} v_x^2 \text, \label{eq:F_x}
\end{align}
where $C_a \Nam{F_{a}}$ is a simple drivetrain, \Nam{$C_a$ is the motor constant}, $C_{r0}$ is the rolling resistance, and $C_{r2}v_x^2$ is the drag resistance.
\Nam{The rolling resistance coefficient $C_{r0}$ primarily captures mechanical and drivetrain losses, while the drag coefficient $C_{r2}$ depends on vehicle geometry and aerodynamic properties. 
Hence, both are treated as constant parameters and adopted from \cite{liniger2015optimization}. In contrast, road surface variations are primarily captured by the Pacejka parameters ($B$, $C$, $D$), which vary with the road surface condition \cite{pacejka2005tire}.}
We denote the vector of road surface parameters as
    $\etabold = \mm{D_r,\! C_r, \! B_r,\! D_f,\! C_f,\! B_f, \!C_a}^\top$ \text.
The dynamic vehicle model \eqref{eq:dyna_model_1} can then be written as depending on $\etabold$ as
\begin{math}
    \dot{\mathbf{x}} = f(\mathbf{x}, \mathbf{u};\etabold) 
\end{math},
which is discretized with a constant time step $\Delta t > 0$ to obtain the following discrete-time dynamics to be used hereafter
\begin{equation}
    \mathbf{x}_{k+1} = \bfx_k + g(\bfx_k, \bfu_k;\etabold) \text,  \label{eq:dyna_ODE}
\end{equation}
where
$g(\bfx_k, \bfu_k;\etabold) = \int_{0}^{\Delta t}f(\mathbf{x}, \mathbf{u};\etabold) \diff t$,
$k$ is the time-step index, $t_k$ is the corresponding continuous time, $\mathbf{x}_k = \mathbf{x}(t_k)$, and $\mathbf{u}_k = \mathbf{u}(t_k)$.
\Nam{Usually, the parameter $\etabold$ of the predictive model is assumed fixed \cite{liniger2015optimization, gomes2024learning}. }
However, in practice, due to uncertainties and varying road conditions, $\etabold$ must be updated to accurately reflect the vehicle's dynamics.


%% file: MPCC.tex
\vspace{-12pt}
\subsection{Model predictive contouring control}
\label{sec:Diff-MPCC}
This paper follows the MPCC formulation presented in \cite{liniger2015optimization, kabzan2019learning}.
The reference path of an MPCC controller is parameterized by a progress variable $\theta \in [0,\theta_{\max}]$,
representing a parametric curve in the global $\mathrm{X}$--$\mathrm{Y}$ frame, with coordinates given by the functions $X_c(\theta)$ and $Y_c(\theta)$. 
In this paper, we assume that $X_c(\theta)$ and $Y_c(\theta)$ describe the racetrack centerline, obtained via interpolation between successive track waypoints.
The vehicle completes the track as $\theta$ increases from 0 to $\theta_\textrm{max}$.
 At the current position $[X_k,~Y_k]$, $\theta_k$ is determined as
\begin{equation}
    \theta_k = \argmin_{\theta\in [0,\;\theta_{\max}]}~ \Vert X_k - X_c(\theta)\Vert_2^2 + \Vert Y_k - Y_c(\theta)\Vert_2^2  \text. \label{eq:cal_theta}
\end{equation}
The lag error $e_l$ and contour error $e_c$ {are defined based on the} position $(X, Y)$ and progress $\theta$ {as}
\begin{align}
    e_c(\mathbf{x}, \theta) &=  \sin(\Phi(\theta))(X \!-\! X_c(\theta)) \!-\! \cos(\Phi(\theta))(Y \!-\! Y_c(\theta)) \text, \notag 
    \\
    e_l(\mathbf{x}, \theta) &=  \!-\! \cos(\Phi(\theta))(X \!-\! X_c(\theta))  \!-\! \sin(\Phi(\theta))(Y \!-\! Y_c(\theta))\text, \notag 
\end{align}
where $\Phi(\theta) = \arctan\big(\nabla Y_c(\theta)/\nabla 
X_c(\theta)\big)$ {is the tangent angle of the reference path, and}
$X_c(\theta_k)$ and $Y_c(\theta_k)$ {denote} the position of the center {line} at $\theta_k$ {obtained}  from  \eqref{eq:cal_theta}. 
{As i}n \cite{kabzan2019learning}, a virtual control input $\vartheta$ is introduced as the progress rate of $\theta$ along the track.
The MPCC optimization problem is then formulated as
\begin{subequations}
\label{eq:MPCC}
\begin{align}
    \min  &\; J(\bfx, \bfu , \thetabold , \boldsymbol{\vartheta} ; \bfq)
    \notag \\
    \text{s.t.}~&{\bfx}_{0|k} = \bfx_{k} \text, 
        \label{eq:MPCC-0}
    \\
    &{\bfx}_{i+1|k} = \bfx_{i|k} + g(\bfx_{i|k}, \bfu_{i|k};\etabold)\text,
        \label{eq:MPCC-1}
    \\
    & \theta_{i+1|k} = \theta_{i|k} + \vartheta_{i|k} \text, 
        \label{eq:MPCC-2}
    \\
    &A_x {\bfx}_{i|k} \preceq \bfb_x \text, 
    \label{eq:MPCC-3}
    \\
    &A_u \bfu_{i|k} \preceq \bfb_u\text,
    \label{eq:MPCC-4}
    \\
    & 0 \leq \theta_{i|k} \leq \theta_{\max}\text,
    \label{eq:MPCC-5}
\end{align}
\end{subequations}
where $T_h$ is the control horizon,
$\bfx  = \col(\bfx_{1|k},\dots, \bfx_{T_h|k})$,
$\bfu  = \col(\bfu_{0|k},\dots, \bfu_{T_h-1|k})$,
$\thetabold  = \mm{\theta_{1|k},\dots, 
\theta_{T_h|k}}^\top$,
$\boldsymbol{\vartheta}  = \mm{\vartheta_{1|k},\dots, 
\vartheta_{T_h|k}}^\top$
with
$\theta_{0|k} = \theta_k$ obtained from \eqref{eq:cal_theta},
\eqref{eq:MPCC-3}, and \eqref{eq:MPCC-4} are the state and control constraints, respectively.
The objective function of MPCC is formulated to increase the virtual control input $\boldsymbol{\vartheta}$, which drives the racing progress forward, while reducing the contouring and lag errors as
\begin{multline*}
J(\bfx , \bfu , \thetabold , \boldsymbol{\vartheta} ; \bfq) =
\\
\sum_{i = 0}^{T_h-1} \bigl( q_c e_c^2(\Nam{\bfx_{i|k}, \theta_{i|k}}) + q_l e_l^2(\Nam{\bfx_{i|k}},  \theta_{i|k})
- q_v \vartheta_{i|k} + \bfu_{i|k}^\top R \bfu_{i|k} \bigr) \text,
\end{multline*}
in which $\mathbf{q} = \mm{q_c, q_l, q_v}^\top$, \, $q_c,q_l,q_v \in \mathbb{R}_{>0}$ are the MPCC objective weights, and $R = R^\top \in \bbR^{2\times 2}$ is a positive definite matrix.


The first element of the optimal input sequence, $\bfu_{0|k}$, is applied to the vehicle at time step $k$.
For simplicity, we denote 
$\mm{\bfu_k, \vartheta_k}^\top = \mm{\bfu_{0|k}, \vartheta_{0|k}}^\top = \MPCC(\bfx_k;\bfq, \etabold)$ where $\mm{\bfu_k, \vartheta_k}^\top \in \bbR^3$.

\textit{Our goal is to develop a real-time MPCC-based control framework for autonomous racing cars that can improve racing performance compared to conventional MPCC while maintaining safety under varying road-surface conditions, through online adaptation of the Pacejka parameters and automatic tuning of the MPCC objective weights.}

Towards this goal, we present in the next section our framework: Adaptive Differentiable MPCC (AD-MPCC).

%% file: sec_method_2.tex
\section{Adaptive Differentiable MPCC}
\label{sec:AD-MPCC}

The AD-MPCC framework comprises two main components, illustrated in Fig.~\ref{fig:adap_MPCC} and described in this section.
The first component updates the varying Pacejka parameters online, based on the MHE approach.
The second component optimally adapts the MPCC objective weights in real time by leveraging advanced differentiable programming.

\subsection{Online Pacejka parameter estimation}

%% file: adaptation.tex
\label{subsec:online_MHE}
This paper employs MHE \cite{kebbati2025learning} to estimate $\etabold$ from the rolling dataset $\mathcal{D}_{k} = \{\mathbf{x}_{i+1}, \mathbf{u}_i\}_{i=k-T, \dots, k-1}$, where
$k$ is the current time step and $T\in \bbN_{>0}$ is the window horizon ($k> T$).
On multi-surface tracks, the dynamic model must adapt quickly to changing tire-road characteristics by prioritizing recent data over older observations. Hence, we introduce exponentially decaying weights into the MHE objective. 
Additionally, we regularize toward nominal Pacejka parameters $\etabold_{gt}$ obtained from the regular road of the racetrack.
Consequently, at time step $k$, we estimate the parameters $\etabold$ from the dataset $\mathcal{D}_{k}$ via the optimization problem
\begin{align}
   \etabold_k =& \argmin_{\etabold \in \Gamma} \sum_{\tau=1}^{T-1} c^\tau \Vert \Delta \bfx_{k-\tau} \!-\!  g(\bfx_{k-\tau}, \bfu_{k-\tau}; \etabold) \Vert_2^2  \notag \\&~~~~~~~~~~~~~~~~~~~~~~~~~~~~~~~~~~~~+  \frac{\sigma}{2} \Vert \etabold - \etabold_{gt}\Vert_2^2, \label{eq:uncstr-adapt}
\end{align}
where $\hat \bfx_{k-\tau}$ is the {predicted} value of $\bfx_{k-\tau}$, {$\Gamma$ is the feasible set for $\etabold$}, $c$ is the {exponential} decay factor ($c < 1$) and $\Delta \bfx_{k-\tau} = \bfx_{k-\tau+1} \!-\! \bfx_{k-\tau}$. 
The optimization problem \eqref{eq:uncstr-adapt} has a differentiable cost and can be {efficiently} solved {via} first-order optimization methods \cite{beck2017first}. 
The parameter $\boldsymbol{\eta}$ is updated at each time step. We denote by $\boldsymbol{\eta}_k$ and $\mathbf{q}_k$ the updated value of $\boldsymbol{\eta}$ and the value of $\mathbf{q}$ at time step $k$.

%% file: diffMPCC.tex
\subsection{Differentiable MPCC (Diff-MPCC)}
This section develops a weight adaptation mechanism using differentiable programming to determine $\mathbf{q}_k$ at each time step. Differentiable MPC \cite{11373898} has been shown to improve control performance by optimizing the MPC weights. However, extending differentiable programming to MPCC is nontrivial. In particular, MPCC involves three coupled objective weights, $q_c$, $q_l$, and $q_v$, each associated with a distinct control objective. Also, the reference path $(X_c(\theta), Y_c(\theta))$ depends on the optimized progress variable $\theta$.

To evaluate the performance of $\MPCC$ over a long horizon $H \gg T_h$, 
let us consider the performance optimization problem in the following form
\begin{subequations}
\label{eq:diffMPC}
\begin{align}
    \min_{\bfq}\; &\ell(\bfq; \bfx_k, \etabold_k) =\sum_{i = 0}^{H-1} l_i(\bar\bfx_{i+1|k}, \bar\bfu_{i|k}, \bar\vartheta_{i|k}),
    \label{eq:diffMPC-0}
    \\
    \text{s.t.}~&\bar\bfx_{i+1|k} = \bar\bfx_{i|k} + g(\bar\bfx_{i|k}, \bar\bfu_{i|k};\etabold_k),~ \bar\bfx_{0|k} = \bfx_k,
    \label{eq:diffMPC-1}
    \\
    &\mm{\bar\bfu_{k|i}, \bar\vartheta_{i|k}}^\top = \MPCC(\bar\bfx_{i|k}; \bfq, \etabold_k), \vspace{30pt}
\label{eq:diffMPC-2}
\end{align}
\end{subequations} 

\vspace{2pt} \noindent
where $\ell$ is the performance cost function, $l_i$ is the twice differentiable convex stage cost function, and $\bfx_k$ and $\etabold_k$ denote the current state and estimated parameters at time step $k$. In \eqref{eq:diffMPC}, we optimize the MPCC weights $\bfq$ at time step $k$ by minimizing the performance cost $\ell$, subject to the system dynamics and the MPCC feedback policy. Here, $\bar{\bfx}_{i|k}$, $\bar{\bfu}_{i|k}$, and $\bar{\vartheta}_{i|k}$ denote the predicted state, input, and virtual input associated with \eqref{eq:MPCC}, respectively. Since the prediction horizon $T_h$ of MPCC is limited by computational resources, the cost $\ell$ represents the closed-loop performance of the MPCC policy over a longer horizon $H$.

We assume that $\mathbf{q}$ belongs to the set $\mathcal{Q}$ of admissible parameters, for example, $q_c, q_\ell, q_v > 0$. We also assume that the MPCC policy \eqref{eq:diffMPC-2} is feasible for all $\bfq \in \mathcal{Q}$. Note that the constrained optimization problem \eqref{eq:diffMPC} can be rewritten as an unconstrained optimization problem by substituting \eqref{eq:diffMPC-1} and \eqref{eq:diffMPC-2} into \eqref{eq:diffMPC-0}. In addition, if the gradient $\pdiff{\ell}{\bfq}$ can be computed, $\bfq_k$ can be optimized using the following projected gradient method
\begin{align}
\label{eq:pol-grad}
    \bfq_k^{(j+1)} = \mathrm{Proj}_{\calQ}\left(\bfq_k^{(j)} - \gamma^{(j)} \pdiff{\ell}{\bfq} \right),
\end{align}

\vspace{2pt} \noindent
where 
$\mathrm{Proj}_{\mathcal{Q}}(\bfq)$ is the projection of $\bfq$ onto $\mathcal{Q}$, $j$ is the iteration index of the gradient-based update, and $\gamma^{(j)} > 0$ is the learning rate. By the chain rule, the gradient of $\ell$ with respect to $\bfq$ can be computed as
\begin{subequations}
\label{eq:diff_ell}
\begin{align}
&\pdiff{\ell}{\bfq} =
\sum_{i=0}^{H-1}
\pdiff{\ell}{\bar\bfx_{i+1|k}}
\pdiff{\bar\bfx_{i+1|k}}{\bfq}
+
\pdiff{\ell}{\bar\bfu_{i|k}}
\pdiff{\bar\bfu_{i|k}}{\bfq} + \pdiff{\ell}{ \bar\vartheta_{i|k}} \pdiff{\bar\vartheta_{i|k}}{\bfq},
\label{eq:diff_ell1}
\\
&\pdiff{\bar\bfx_{i+1|k}}{\bfq} = \pdiff{\bar\bfx_{i|k}}{\bfq} + \pdiff{g}{\bar\bfx_{i|k}} \pdiff{\bar\bfx_{i|k}}{\bfq} + \pdiff{g}{\bar\bfu_{i|k}} \pdiff{\bar\bfu_{i|k}}{\bfq},
\label{eq:diff_ell2}
\\
&\mm{\pdiff{\bar\bfu_{i|k}}{\bfq}; \pdiff{\bar\vartheta_{i|k}}{\bfq}} = \pdiff{\pi_q}{\bfq} (\bar\bfx_{i|k};\bfq_k^{(j)}, \etabold_k) \text.
\label{eq:diff_ell3}
\end{align}
\end{subequations} 

\vspace{2pt} \noindent
Computing $\pdiff{\ell}{\bfq}$ via \eqref{eq:diff_ell} requires $H$ iterations from $0$ to $H-1$, starting from $\pdiff{\bar\bfu_{0|k}}{\bfq}$ and $\pdiff{\bar\bfx_{1|k}}{\bfq}$ to $\pdiff{\bar\bfu_{H-1|k}}{\bfq}$ and $\pdiff{\bar\bfx_{H|k}}{\bfq}$. The main difficulty lies in computing $\pdiff{\pi_q}{\bfq}$, since there is no explicit mapping from $\bfq_k$ to the solution of MPCC. In each iteration of the gradient-based update \eqref{eq:pol-grad}, the MPCC optimization problem \eqref{eq:MPCC} must be solved $H$ times to compute $\pdiff{\ell}{\bfq}$ when $\bfq_k^{(j)}$ is updated. Indeed, to compute $\pdiff{\pi_q}{\bfq}(\bar\bfx_{0|k}; \bfq_k^{(j)}, \boldsymbol{\eta}_k)$, \eqref{eq:MPCC} must be solved at $(\bar\bfx_{0|k}, \bfq_k^{(j)}, \boldsymbol{\eta}_k)$ to obtain $\bar\bfu_{0|k}$ and $\bar{\vartheta}_{0|k}$. Then, by \eqref{eq:diffMPC-1}, we can compute $\bar\bfx_{1|k}$. Repeatedly applying this procedure, we obtain 
$\{\bar\bfx_{i|k}\}_{i=1,\dots,H}$, 
$\{\bar\bfu_{i|k}\}_{i=0,\dots,H-1}$, and 
$\{\bar{\vartheta}_{i|k}\}_{i=0,\dots,H-1}$. 


Let $h_{\mathrm{eq}}(\bfz; \etabold_k) = 0$ denote the equality constraints \eqref{eq:MPCC-0}–\eqref{eq:MPCC-2}, where $\bfz = \col(\bfx, \bfu, \boldsymbol{\theta}, \boldsymbol{\vartheta})$. The inequality constraints \eqref{eq:MPCC-3}–\eqref{eq:MPCC-5} are denoted by  $h_{\mathrm{in}}(\bfz) \preceq 0$. A common approach to compute \( \pdiff{\pi_q}{\mathbf{q}} \) is sequential quadratic programming (SQP) \cite{adabagDifferentiableModel2025}, which approximates the nonconvex optimization problem \eqref{eq:MPCC} by a quadratic program through linearization of the dynamics \eqref{eq:MPCC-1} and quadratic approximation of the cost function $J(\bfz;\bfq)= J(\bfx, \bfu, \boldsymbol{\theta}, \boldsymbol{\vartheta};\bfq)$.
Although SQP can accelerate {the} computation of $\pdiff{\pi_q}{\bfq}$, the approximation error can accumulate over the long horizon $H$, potentially violating the dynamics constraint \eqref{eq:diffMPC-1} {and can lead to unsafe control actions.}
Moreover, the existence of \( \pderiv{\pi_q}{\mathbf{q}} \) cannot be guaranteed when the parameter $\mathbf{q}_k$ is updated according to \eqref{eq:pol-grad}, particularly when the cost function $J(\mathbf{z};\mathbf{q})$ is nonconvex and the optimization problem may admit multiple local minima.

Instead, we approximately compute \( \pdiff{\pi_q}{\mathbf{q}} \) using nonlinear programming methods by applying the implicit function theorem (IFT) \cite{blondel2024elements} to the optimality conditions. Specifically, consider the following optimization problem at the $j$-th iteration of the policy update
\begin{align}
    (\calP_j):\;\min_{\bfz}\; \bar J^{(j)}(\bfz; \bfq_k^{(j)}, \etabold_k) 
    \label{eq:IPopt-diff}
    \text{~s.t.~}\;h_\mathrm{eq}(\bfz;\etabold_k) = 0,
\end{align}
where
\begin{multline*}
\bar J^{(j)}(\bfz; \bfq_k^{(j)}, \etabold_k) = J(\bfz;\bfq_k^{(j)}) - \epsilon_1 
\sum_i^{n_h} \log(-h_{\text{in},i}(\bfz)) \\
+ \frac{\epsilon_2}{2} \Vert \bfz - \bfz^{(j-1)} \Vert_2^2\text,        
\end{multline*}

\vspace{1pt} \noindent
in which
$h_{\mathrm{in},i}$ are log barrier functions,
$n_h$ is the number of inequalities,
$\epsilon_1 > 0$ is the barrier parameter,
$\epsilon_2 > 0$, and
$\bfz^{(j-1)} = \col(\bfx^{(j-1)}, \bfu^{(j-1)}, \thetabold^{(j-1)}, \boldsymbol{\vartheta}^{(j-1)})$ is a (local) minimum of $\calP_{j-1}$.
Let us define $h_\text{in}(\bfz) = \mm{h_{\mathrm{in},1}(\bfz), h_{\mathrm{in},2}(\bfz), \dots, h_{\mathrm{in},n_h}(\bfz)}^\top$.
Note that the solutions of $\calP_j$ approach those of \eqref{eq:MPCC} as $\epsilon_1, \epsilon_2 \to 0$ when $\bfq = \bfq_k^{(j)}, \etabold = \etabold_k$.
Thus, by choosing small enough $\epsilon_1$ and $\epsilon_2$, instead of computing $\pdiff{\pi_q}{\bfq} (\bfx_{i|k};\bfq_k^{(j)}, \etabold_k)$, we can approximate it by 
$\pdiff{\bar{\pi}_q}{\bfq} (\bfx_{i|k};\bfq_k^{(j)}, \etabold_k)$ where $\bar{\pi}_q(\bfx_{i|k};\bfq_k^{(j)}, \etabold_k)$ is the approximated MPCC controller by $\calP_j$.

Solving problem $\calP_j$ is with using $\bfz^{(j-1)}$, obtained from $\calP_{j-1}$, as the warm start.
The Lagrangian of \eqref{eq:IPopt-diff} is 
$L(\bfz, \lambdabold, \betabold;\bfq_k^{(j)}, \etabold_k) =  \bar J^{(j)}(\bfz; \bfq_k^{(j)}, \etabold_k) + \lambdabold^\top h_\mathrm{eq}(\bfz;\etabold_k) \!+\! \betabold^\top h_\mathrm{in}(\bfz)$, and
the first order optimality conditions of \eqref{eq:IPopt-diff} are given by
\begin{subequations}
\label{eq:KKT-ip}
\begin{align}
     0 &= \nabla_{\bfz} L(\bfz, \lambdabold, \betabold;\bfq_k^{(j)}, \etabold_k),
     \label{eq:KKT-ip-1}
    \\
    0 &= h_\mathrm{eq}(\bfz;\etabold_k),
         \label{eq:KKT-ip-2}
    \\
    0 &= \diag(\betabold) h_\mathrm{in}(\bfz;\etabold_k) + \epsilon_1 \mm{1,\dots,1}^\top,
         \label{eq:KKT-ip-3}
    \\
    0 &\prec \betabold.
\end{align}
\end{subequations}
By the IFM, letting $\bfw = \col(\bfz, \lambdabold, \betabold)$, we can compute 
\begin{align}
     \pderiv{\bar{\pi}_q}{\bfq}^\top  = \pderiv{\bfw}{\bfq}^\top 
\pderiv{\bar{\pi}_q}{\bfw}^\top
\underset{\mathrm{IFM}}{=\;\;\;} -  \pderiv{\boldsymbol{\kappa}}{\bfq}^\top 
\underbrace{\left(\pderiv{\boldsymbol{\kappa}}{\bfw}^\top\right)^{\!-1} \!C}_{\bfv} ,
     \label{eq:diff-u-p1}
\end{align}
where 
$\pderiv{\bar{\pi}_q}{\bfw} = C^\top = 
\mm{1, 0, \dots, 0, \dots, 0; 
    0, 1,\dots, 0, \dots, 0; 
    0, 0,\dots, 1, \dots, 0}$,
$\boldsymbol{\kappa}$ is the vector function whose elements are gathered from the right-hand side of \eqref{eq:KKT-ip-1}-\eqref{eq:KKT-ip-3}, and
\begin{align*}
    \pderiv{\boldsymbol{\kappa}}{\bfw} =
    \begin{bmatrix}
    \nabla_{\bfz\bfz}^2 L &\!\!\!\!\!\!\partial_\bfz^\top h_\mathrm{eq} (\bfz^{(j)}; \etabold_k)\!\!\!\!\!\! &\partial_\bfz^\top h_\mathrm{in}(\bfz^{(j)}) \!\\
    \partial_\bfz h_\mathrm{eq}(\bfz^{(j)};\etabold_k) &0 &0\!\\
    \diag(\betabold) \partial_\bfz h_\mathrm{in}(\bfz^{(j)}) &0 &\diag(h_\mathrm{in}(\bfz^{(j)})\!)\!
    \end{bmatrix}\!\!.
\end{align*}    
In \eqref{eq:diff-u-p1}, $\bfv$ is computed effectively by exploiting the sparsity of $\pderiv{\boldsymbol{\kappa}}{\bfw}$ to solve
$\pderiv{\boldsymbol{\kappa}}{\bfw}^\top \bfv = C$.
Algorithm~\ref{alg:diffMPC} summarizes the gradient-based weights update for given $\etabold_k$ and $\bfx_k$.

\begin{remark}
Several methods exist to compute $\pderiv{\pi_\bfq}{\bfq}$, including differentiable nonlinear MPC \cite{frey2025differentiable, zuliani2025differentiable}, linearized MPC \cite{adabagDifferentiableModel2025}, and JAXOPT \cite{blondel2022efficient}. In the proposed differentiable MPCC (Diff-MPCC) framework, the nonsingularity of the Hessian matrix $\nabla_{\mathbf{z}\mathbf{z}}^2 L$ is enforced by incorporating convex terms into \eqref{eq:IPopt-diff}, namely barrier and regularization functions. Furthermore, since $h_\mathrm{in}(\mathbf{z}) < 0$, the KKT matrix $\pderiv{\boldsymbol\kappa}{\bfw}$ is invertible if the Jacobian $\partial_{\mathbf{z}} h_\mathrm{eq}(\mathbf{z};\etabold_k)$ has full row rank and the Hessian matrix $\nabla^2_{\mathbf{z}\mathbf{z}} L$ is nonsingular.
\end{remark}

\begin{algorithm}[!t]
\caption{Gradient-based MPCC weights update}
\label{alg:diffMPC}
\begin{algorithmic}
\Require Set of MPCC weights $\calQ$, state $\bfx_k$ and parameter $\etabold_k$ at step $k$, loss performance $\ell$,  initial weights $\bfq_k^{(0)}$, and number of iteration $n_\mathrm{iter}$.
\Ensure MPCC weights $\bfq_k^{(j)} \in \calQ$
\For{$j = 1,2,\dots, n_\mathrm{iter}$}
\State Set $\bar\bfx_{0|k} = \bfx_k$
\For{$i = 0,1,\dots, H-1$}
\State Compute $\bfx^{(j)}, \bfu^{(j)}, \boldsymbol\theta^{(j)}, \boldsymbol\vartheta^{(j)}$ in \eqref{eq:IPopt-diff} 
\State Set $\bar\bfu_{i|k} = \bfu_{0|k}^{(j)}$ and $\bar\vartheta_{i|k} = \vartheta_{0|k}^{(j)}$
\State Compute $\bar\bfx_{i+1|k}$ from $\bar\bfx_{i|k}$ and $\bar\bfu_{i|k}$ in \eqref{eq:diffMPC-1}
\State Compute $\pderiv{\boldsymbol{\kappa}}{\bfw}$ at  $\bfx^{(j)}, \bfu^{(j)}, \boldsymbol\theta^{(j)}, \boldsymbol\vartheta^{(j)}$
\State Compute $\pdiff{\bar\bfu_{i|k}}{\bfq}$ and $\pdiff{\bar\vartheta_{i|k}}{\bfq}$ in \eqref{eq:diff_ell3} by \eqref{eq:diff-u-p1}
\State Compute $\pdiff{\bar\bfx_{i+1|k}}{\bfq}$ in \eqref{eq:diff_ell2}
\EndFor
 \State Compute $\pdiff{\ell}{\bfq}$ in \eqref{eq:diff_ell1}
\State Update MPCC weights $\bfq_k^{(j+1)}$ in \eqref{eq:pol-grad}
\EndFor
\end{algorithmic}
\end{algorithm}

\begin{remark}  
Unlike differentiable MPC, differentiable MPCC requires higher-order derivatives of the reference path, namely $\partial_\theta X_c(\theta)$, $\partial_\theta Y_c(\theta)$, $\nabla^2_{\theta\theta} X_c(\theta)$, and $\nabla^2_{\theta\theta} Y_c(\theta)$, as they explicitly appear in $\nabla^2_{\mathbf{z}\mathbf{z}} L$. Hence, the reference path $(X_c(\theta), Y_c(\theta))$ must be at least twice continuously differentiable to ensure well-defined and continuous second-order derivatives.
\end{remark}



\subsection{Online adaptation of MPCC weights}
\label{subsec:MPCC-online}


At each time step $k$, changes in the Pacejka parameters require re-running
Algorithm~\ref{alg:diffMPC} to compute new optimal MPCC weights $\mathbf{q}$, which may exceed real-time computation budgets.
To address this issue, we approximate the mapping from $(\mathbf{x}_k, \etabold_k)$ to the optimal weights via a supervised learning model trained offline.

The training data is generated using a racing vehicle simulated under various road-surface conditions. The dataset consists of state vectors $\mathbf{x}_k^l$ and corresponding tire parameter vectors $\boldsymbol{\eta}_k^l$, where $k$ denotes the time index and $l$ denotes the road-surface condition.
For each road type $l$, the constant Pacejka parameters $\boldsymbol{\eta}^l$ are computed using the semi-empirical formulation described in \cite[Chapter 4]{pacejka2005tire}.
Given $\boldsymbol{\eta}^l$, a standard MPCC controller with the dynamic model parameterized by $\boldsymbol{\eta}^l$ is employed to generate the training data. 
We define the dataset $\mathcal{S} = \{\{(\mathbf{x}_k^l, \etabold^l), \mathbf{q}^l_k\}_{k = 0, \dots, N_l}\}_{l = 1, \dots, L}$, where $N_l$ is the number of data points collected on road $l$, and $L$ is the total number of considered roads.
The label $\mathbf{q}_{k}^l$ is obtained by Algorithm \ref{alg:diffMPC} with given $(\mathbf{x}_k^l, \etabold^l)$.

\begin{figure}[t]
    \centering
    \includegraphics[trim={1.0cm 0 0.75cm 0},clip, width=1.0\linewidth]{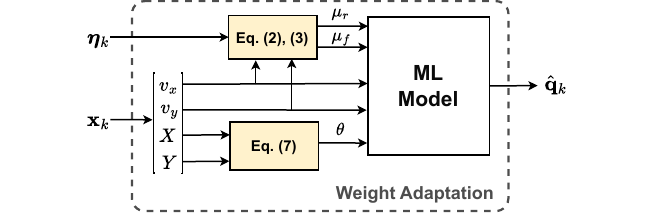}
    \vspace{-18pt}
    \caption{
    Diagram of Pacejka-informed machine learning (PaIML) model for MPCC weights online adaptation.
    } 
    \label{fig:diff_look}
\end{figure}

\begin{figure*}[t]
    \centering
    \includegraphics[trim={0.25cm 0.25cm 0.cm 0.25cm},clip,width= 1.0\linewidth]{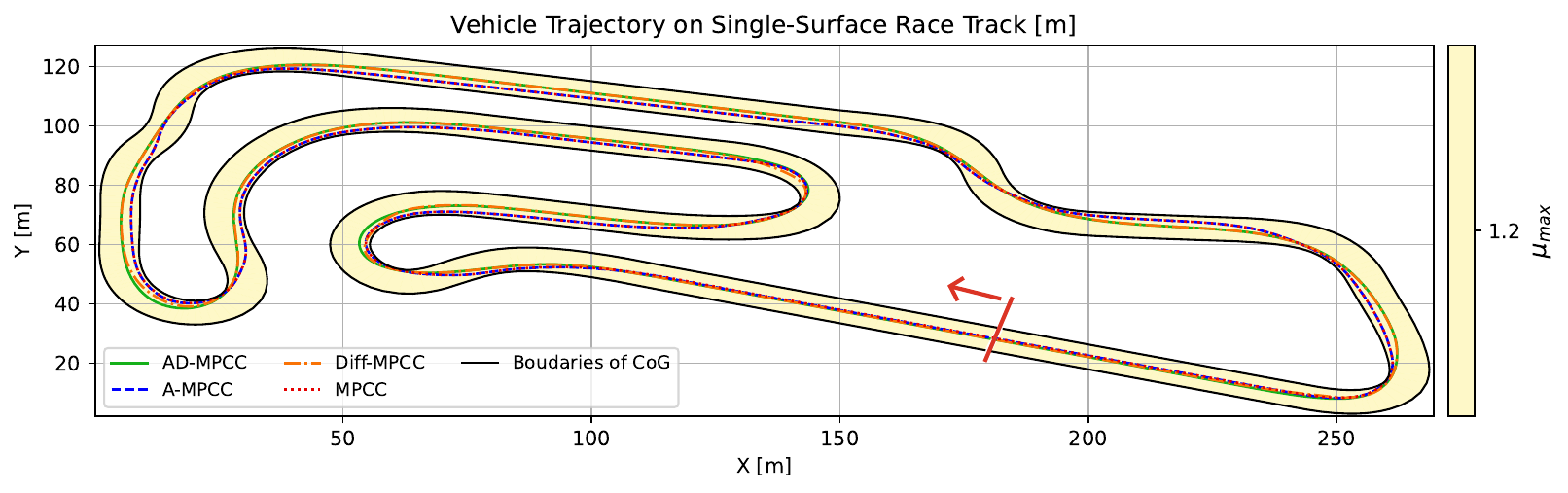}
    \vspace{-20pt}
    \caption{The trajectories of the four MPCC controllers during the ninth lap. The road surface is uniform with $\mu_{\max} = 1.2$. All methods complete 10 laps.
    }
    \label{fig:traj}
    \vspace{-10pt}
\end{figure*}
\begin{figure}[t]
    \vspace{-5pt}
    \centering
    \includegraphics[trim={0.0cm 0.0cm 0.0cm 5.5cm},clip, width=0.92\linewidth]{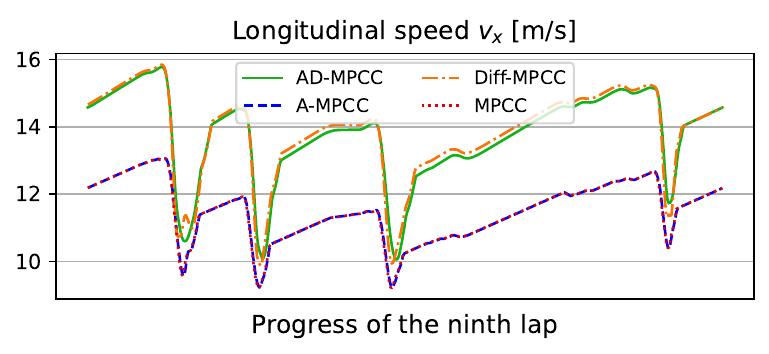}
    \vspace{-10pt}
    \caption{Longitudinal velocity during the ninth lap in the single-surface scenario as the vehicle completes the racetrack from left to right.}
    \label{fig:velo}
\end{figure}

\begin{figure}[t]
    \centering
    \includegraphics[trim={0.65cm 0.0cm 0.0cm 5.75cm},clip, width=0.92\linewidth]{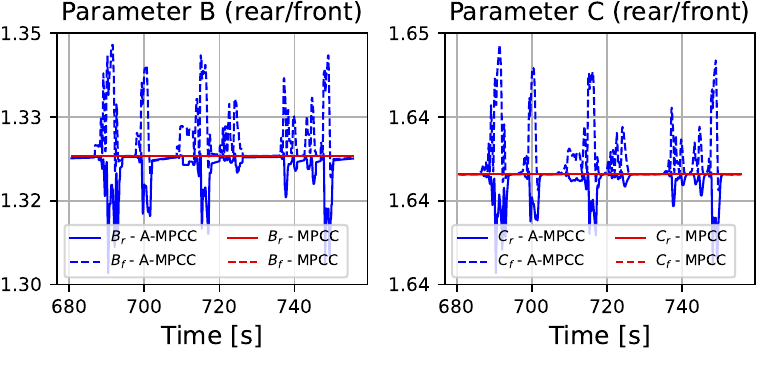}
    \vspace{-10pt}
    \caption{Online-estimated and fixed Pacejka parameters (results for sample B and C only) of the ninth lap in single-surface road scenario. 
    }
    \label{fig:param}
\end{figure}

In practice, obtaining exact tire parameters in real-world environments is generally difficult due to modeling uncertainties, and constructing test conditions with different road surfaces for data collection can be costly.
Hence, data can be collected using high-fidelity simulators, such as F1TENTH-Gym \cite{o2020f1tenth} for autonomous racing or CARLA \cite{dosovitskiy2017carla} for general autonomous driving, to generate representative datasets prior to real-world deployment.

At each time step $k$, using the full input $\mm{\mathbf{x}_k, \etabold_k}^\top$ results in a 14-dimensional feature vector, which increases inference time and makes the ML model more complex.
To deal with this issue, we propose Pacejka-informed machine learning (PaIML) in Fig. \ref{fig:diff_look}, which leverages the physical structure of the Pacejka's magic formula together with an ML model to infer the optimal weights from lower-dimensional inputs.
PaIML is also considered as a physics-informed ML technique \cite{NamAccess2026, nguyen2026structurestabilitypreservinglearningporthamiltonian}.
In particular, $\etabold$ mainly influences the lateral tire forces through the friction coefficients $\mu_r$ and $\mu_f$, which are computed from $\etabold$ via \eqref{eq:mu_r_y} and \eqref{eq:mu_f_y}.
Thus, we replace the input $\etabold$ with the reduced representation consisting of $\mu_r$ and $\mu_f$ for the inference model.
Moreover, the vehicle position $(X, Y)$ encodes track curvature through the progress variable $\theta$ obtained from \eqref{eq:cal_theta}. 
Together with the longitudinal and lateral velocities $v_x$ and $v_y$, these variables capture the dominant information describing the vehicle state. 
This results in the 5-dimensional input vector $\mm{\mu_r, \mu_f, v_x, v_y, \theta}^\top$.

Consequently, PaIML learns the mapping from the five inputs to the optimal MPCC weights $\hat{\mathbf{q}}$, as illustrated in Fig.~\ref{fig:diff_look}. 
Since the regression task is low-dimensional (five inputs and three outputs) and a large dataset can be generated offline, we employ XGBoost \cite{chen2016xgboost} for the ML component of PaIML. 
Tree-based ensemble methods are well suited for structured low-dimensional data and provide built-in regularization, resulting in robust performance without extensive hyperparameter tuning.

\begin{table}[t]
\centering
\caption{Results of single-surface case for 10 laps.}
\vspace{-5pt}
\label{tab:7_round}
\setlength{\tabcolsep}{2.5pt}
\renewcommand{\arraystretch}{1.15}
\begin{tabular}{|l|c|c|c|c|}
\hline
 & MPCC~\cite{liniger2015optimization} & A-MPCC & Diff-MPCC & AD-MPCC \\ \hline

Avg. lap Time [\unit{\second}] 
& 75.57 
& 75.62 
& \textbf{64.08} 
& 64.89 \\ \hline

Avg.  $v_x$ [\unit{\meter \per \second}] 
& 11.383 
& 11.41
& \textbf{13.591} 
& 13.501 \\ \hline

Avg. lateral offset [\unit{\meter}] 
& 1.953 
& 1.815 
& 1.040 
& \textbf{0.926} \\ \hline

Avg. comp. time [\unit{\milli\second}] 
& \textbf{20.45} 
& 21.21 
& 23.56 
& 24.18 \\ \hline
\end{tabular}
\end{table}

%% file: sec_results.tex

\vspace{0pt}
\section{Simulation Results}
\vspace{-2pt}
\label{sec:results}
\subsection{Scenario setup}
To evaluate the proposed AD-MPCC, 
we compare it against three different MPCC schemes: 

\begin{itemize}
\item \textbf{MPCC}: A baseline controller that employs a time-invariant predictive model with fixed Pacejka parameters $\boldsymbol{\eta} = \boldsymbol{\eta}_{gt}$, where $\boldsymbol{\eta}_{gt}$ is obtained from the semi-empirical formulation  \cite{pacejka2005tire} with $\mu_{\max} = 1.2$.

\item \textbf{A-MPCC}: Ablation which only adopts the online parameter estimation from \eqref{eq:uncstr-adapt} for the predictive model in the MPCC, instead of a time-invariant model.

\item {\textbf{Diff-MPCC}: Ablation which only} updates the MPCC weights via {the trained} PaIML model given the states $\mathbf{x}$ and fixed parameters $\etabold_{gt}$.

\end{itemize}

To train the PaIML model for the Diff-MPCC and AD-MPCC methods, data are collected by running a vehicle with the standard MPCC controller using $\etabold$ across different single-surface road conditions, as presented in Subsection~\ref{subsec:MPCC-online}.
Therein, $\mu_{\textrm{max}}$ varies from $0.6$ to $1.2$ in increments of $0.1$. 
The vehicle is initialized with $v_x = \qty{10}{\meter\per\second}$ and completes one lap for each condition, resulting in a dataset of 15{,}992 samples used to train the PaIML model.

\Nam{
To prioritize vehicle progress speed while maintaining safety, we select the stage cost $\ell = 0.01e_c^2 + e_l^2 - 400\vartheta$, thereby assigning a significantly larger weight to $\vartheta$ relative to $e_c$ and $e_l$. Additionally, the prediction horizon is set to $H = 100$, which corresponds to a 5-second rollout in the outer loop of the MPCC.
Algorithm~\ref{alg:diffMPC} is then used with the specified $\ell$ and $H$ to generate the output labels for the PaIML model. 
The PaIML model is implemented using XGBoost with 100 gradient-boosted trees and a maximum depth of 6.
For a fair comparison, the objective weights of MPCC and A-MPCC are fixed at $q_c = 1$, $q_l = 1000$, and $q_v = 250$. 
The MHE parameters are $(T,c,\sigma)=(10,0.8,1.0)$ for the single-surface case and $(5,0.6,0.2)$ for the multi-surface case.}

\begin{figure*}[t]
    \centering
    \includegraphics[trim={0.25cm 0.25cm 0 0.25cm},clip,width=1.0\linewidth]{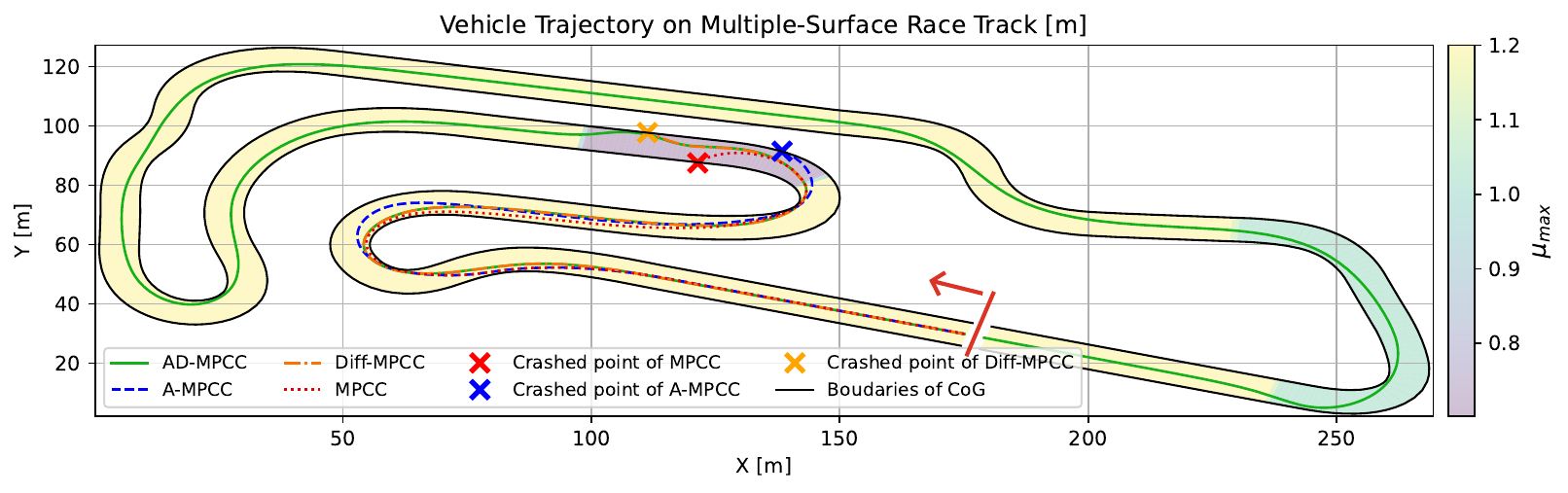}
    \vspace{-20pt}
    \caption{The trajectories of the racing with the four MPCC controllers.  The road consists of multiple surfaces, with the $\mu_{\max}$ varying from 0.7 to 1.2.}
    \label{fig:traj_multi}
    \vspace{-12pt}
\end{figure*}

\begin{figure}[t]
    \vspace{-5pt}
    \centering
    \includegraphics[trim={0.0cm 0.0cm 0.0cm 5.5cm},clip,width=0.92\linewidth]{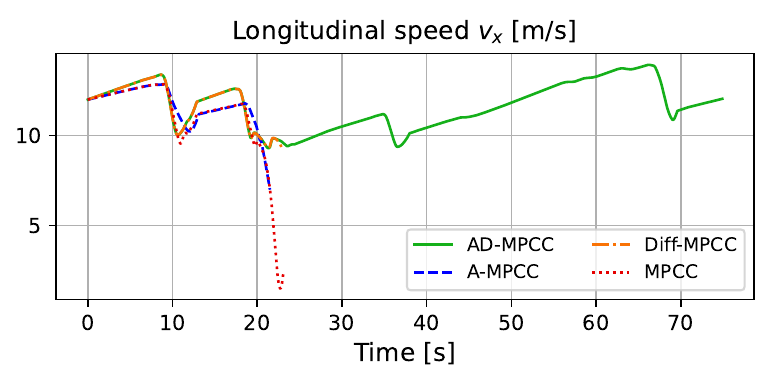}
    \vspace{-12pt}
    \caption{Longitudinal velocity of the multiple-surface case. Only AD-MPCC can complete the racetrack.}
    \label{fig:velo_multi}
\end{figure}
\begin{figure}[t]
    \centering
    \includegraphics[trim={0.65cm 0.0cm 0.0cm 5.75cm},clip, width=0.92\linewidth]{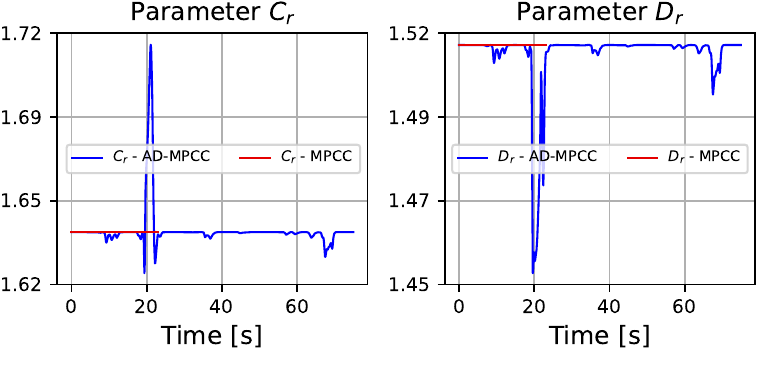}
    \vspace{-12pt}
    \caption{Online-estimated and fixed Pacejka parameters (results for sample B and C only) in multiple-surface road scenario.}
    \label{fig:param_multi}
\end{figure}

\begin{table}[t]
\centering
\caption{Results of multi-surface case for 1 lap.}
\vspace{-5pt}
\label{tab:multi_surface}
\setlength{\tabcolsep}{2.5pt}
\renewcommand{\arraystretch}{1.15}
\begin{tabular}{|l|c|c|c|c|}
\hline
 & MPCC~\cite{liniger2015optimization} & A-MPCC & Diff-MPCC & AD-MPCC \\ \hline

Avg. lap time [\unit{\second}] 
& \textit{crashed} 
& \textit{crashed} 
& \textit{crashed} 
& 74.9 \\ \hline

Avg. $v_x$ [\unit{\meter \per \second}] 
& \textit{crashed} 
& \textit{crashed} 
& \textit{crashed} 
& 11.65 \\ \hline

Avg. lateral offset [\unit{\meter}] 
& \textit{crashed} 
& \textit{crashed} 
& \textit{crashed} 
    & 1.069 \\ \hline

Avg. comp. time [\unit{\milli\second}] 
& \textit{crashed} 
& \textit{crashed} 
& \textit{crashed} 
& 14.91 \\ \hline
\end{tabular}

\vspace{2pt}
\end{table}


The simulation environment is configured with a constant sampling time of \Nam{$\Delta t = \qty{50}{\milli\second}$} and executed on \Nam{a desktop PC equipped with an Intel\textsuperscript{\textregistered} Core\texttrademark{} i7-14700KF processor (20 cores, 28 logical processors, 3.4~GHz), and 32~GB of DDR5 memory}.
The MPCC prediction horizon is set to $T_h = 20$, corresponding to a \Nam{\qty{1}{\second}} look-ahead. The resulting nonlinear optimization problem is modeled in \textsc{CasADi}~\cite{andersson2019casadi} {and solved by} \textsc{IPOPT}~\cite{wachter2006implementation} on the CPU.
Besides, to improve solving time, we may use commercial optimization solvers, such as \textsc{FORCES Pro} \cite{Zanelli02012020} and KNITRO \cite{byrd2006knitro}.

For evaluation, the Oschersleben racetrack from the F1TENTH-Gym environment~\cite{o2020f1tenth} is used. Its compact layout and frequent curves make it challenging for autonomous racing due to limited recovery space and continuous high-curvature maneuvering. 
Experiments are conducted using a full-scale AMZ autonomous racing vehicle model under two scenarios: a known single-surface racetrack and an unknown multi-surface racetrack. 
The half-track width is $\qty{4}{\meter}$, and a collision is declared if the lateral offset between the vehicle's CoG and the centerline exceeds this limit.
\vspace{-5pt}
\subsection{Single-surface road}

In this scenario, all MPCC-based controllers are evaluated on a known road surface with $\mu_\textrm{max} = 1.2$. 
The vehicle is initialized with $v_x = \qty{10}{\meter\per\second}$ and runs for 10 laps.

From Table~\ref{tab:7_round}, Diff-MPCC and AD-MPCC complete 10 laps faster than the standard MPCC controller. 
Specifically, Diff-MPCC achieves the lowest lap time of $\qty{64.08}{\second}$, compared to $\qty{75.57}{\second}$ for MPCC. 
Moreover, AD-MPCC achieves a similar lap time of $\qty{64.89}{\second}$ while maintaining a smaller mean lateral offset of $\qty{0.926}{\meter}$ compared to $\qty{1.040}{\meter}$ for Diff-MPCC, indicating that AD-MPCC provides a better trade-off between safety and speed. 
A-MPCC is about \qty{0.05}{\second} slower per lap than MPCC but reduces the lateral offset by \qty{0.14}{\meter}.

Consistent with these faster lap times, Diff-MPCC and AD-MPCC exhibit higher mean longitudinal velocities $v_x$, although they require slightly longer average computation time per step than MPCC. The MPCC baseline averages $\qty{20.45}{\milli\second}$, while our methods range from $\qty{21.21}{\milli\second}$ to $\qty{24.18}{\milli\second}$ (smaller than $\Delta t = \qty{50}{\milli\second}$).
Additionally, Diff-MPCC and AD-MPCC have smaller mean lateral offsets (below $\qty{1.05}{\meter}$), which is approximately half of those observed for MPCC and A-MPCC. 
This indicates that the controllers still have margin to further increase speed by prioritizing progress maximization via $q_v$ in \eqref{eq:MPCC}.

Figs.~\ref{fig:traj}, \ref{fig:velo}, and \ref{fig:param} show the vehicle trajectory, longitudinal velocity, and Pacejka parameters for the ninth lap. We report the ninth lap as the vehicle has reached steady-state behavior by this point. Fig.~\ref{fig:velo} shows that Diff-MPCC and AD-MPCC maintain higher velocities than A-MPCC and MPCC for most of the lap.
Fig.~\ref{fig:param} shows that, even on a known surface, the estimated $\etabold$ slightly fluctuates due to the simplified Pacejka model and simulation uncertainties, but remains close to the nominal constant values.
\vspace{-2pt}
\subsection{Multiple-surface road}
This scenario considers a multiple-surface road where $\mu_{\textrm{max}}$ varies from $0.7$ to $1.2$. The vehicle is initialized with a longitudinal velocity of $v_x = \qty{12}{\meter\per\second}$ and runs for 1 lap.

Fig.~\ref{fig:traj_multi} shows that only AD-MPCC successfully completes the lap.
In contrast, MPCC, Diff-MPCC, and A-MPCC all crash at the first slippery segment where $\mu_{\textrm{max}} = 0.75$, indicating that fixed or partially adapted models are insufficient to handle sudden friction changes.
Fig.~\ref{fig:param_multi} illustrates that the Pacejka parameters vary in this region when estimated online. For example, the estimated parameter $D_r$ decreases to approximately $1.45$, compared to $1.55$ for the nominal model, reflecting the reduced tire–road friction.
Table~\ref{tab:multi_surface} and Fig.~\ref{fig:velo_multi} show that AD-MPCC not only maintains stability but also achieves competitive performance, with a mean velocity of $\qty{11.65}{\meter\per\second}$ and an average lateral offset of $\qty{1.069}{\meter}$.
The relatively small lateral offset indicates that AD-MPCC still has margin to further increase speed.

In the single-surface scenario, AD-MPCC achieves the smallest lateral error compared to the other methods. 
This smaller lateral offset contributes to improved safety, which becomes particularly important when racing in unknown road conditions. 
As shown in Fig.~\ref{fig:traj_multi}, AD-MPCC is the only controller safely navigating the slippery road segment. 
Our code is published on \href{https://github.com/nxt-lab/AD-MPCC.git}{https://github.com/nxt-lab/AD-MPCC.git}.

%% file: sec_conclusion.tex

\section{Conclusion}
\label{sec:conclusion}

This paper presented AD-MPCC, a framework that jointly adapts the predictive vehicle model and tunes MPCC objective weights for autonomous racing under multiple road-surface conditions. 
The approach combines a prior-regularized moving-horizon estimator for online Pacejka parameter estimation with a differentiable MPCC formulation for weight adaptation, which is approximated offline using a Pacejka-informed machine learning model for real-time deployment.
Simulation results in the F1TENTH-Gym environment demonstrate improved performance on known tracks and robustness in unknown multiple-surface scenarios. 
Specifically, AD-MPCC achieves a lap time approximately \qty{11}{\second} faster than the standard MPCC on the known single-surface racetrack and is the only method capable of completing the unknown multiple-surface race.
Despite these promising results, the current framework relies on simulator-generated data. 
Future work will focus on real-world deployment on an autonomous racing platform and on providing deeper theoretical analysis to establish the stability and robustness of AD-MPCC.